\documentclass[11pt]{article}

\usepackage{microtype}
\usepackage{graphicx}
\usepackage{booktabs}
\usepackage{microsoft-tech-report}
\usepackage{hyperref}
\usepackage{amsmath}
\usepackage{amsfonts}

\usepackage{amssymb}
\usepackage{bm}
\usepackage{enumitem}
\usepackage{multirow}
\usepackage{xspace}
\usepackage{makecell}
\usepackage{algorithm}
\usepackage{algorithmic}

\hypersetup{
  colorlinks=true,
  linkcolor=microsoftblue,
  citecolor=microsoftblue,
  urlcolor=microsoftblue,
  pdftitle={Latent Spatial Memory for Video World Models},
  pdfauthor={Weijie Wang, Haoyu Zhao, Yifan Yang, Feng Chen, Zeyu Zhang, Yefei He, Zicheng Duan, Donny Y. Chen, Yuqing Yang, Bohan Zhuang},
  pdfsubject={arXiv preprint},
  pdfkeywords={Video Generation, Spatial Memory, 3D-consistent Video Generation, Video World Models}
}

\newcommand{\method}{LSM-World\xspace}

\newcommand{\lsmkeywords}{Video Generation, Spatial Memory, 3D-consistent Video Generation, Video World Models}
\newcommand{\blfootnote}[1]{%
  \begingroup
  \renewcommand\thefootnote{}\footnotetext{#1}%
  \addtocounter{footnote}{-1}%
  \endgroup
}
\techreportshorttitle{Latent Spatial Memory}

\begin{document}
\thispagestyle{empty}
\enlargethispage{1.2cm}

\noindent
\begin{minipage}[c]{0.5\linewidth}
\raggedright
\raisebox{-0.5\height}{\includegraphics[height=0.75cm]{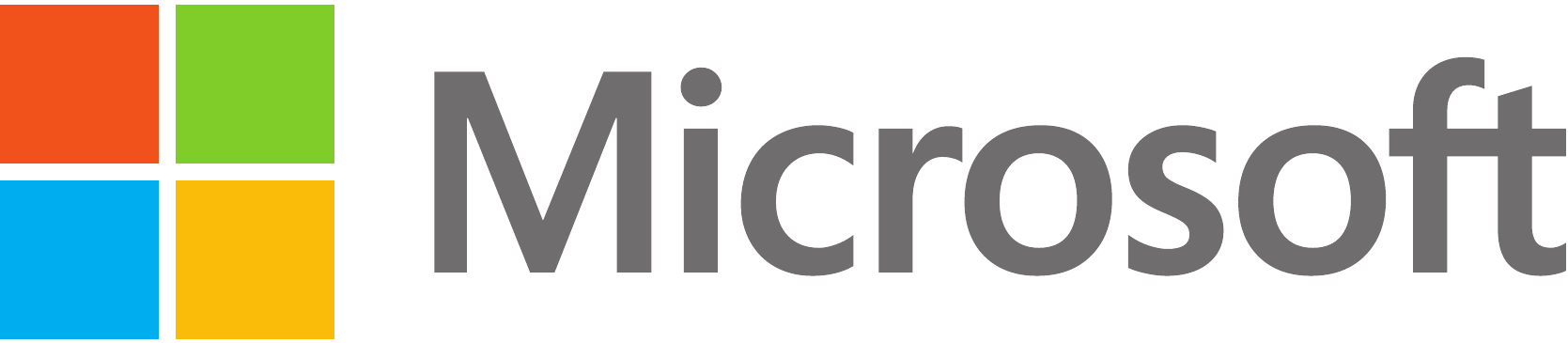}}
\end{minipage}
\begin{minipage}[c]{0.49\linewidth}
\raggedleft
{\microsoftdatefont\small\color{microsoftgray}June, 2026}
\end{minipage}\par
\vspace{0.18em}
\noindent{\color{microsoftline}\rule{\linewidth}{0.8pt}\par}

\vspace{0.42em}
\begin{center}
{{\microsofttitlefont\fontsize{18}{21}\selectfont\color{microsoftdark}
Latent Spatial Memory for Video World Models\par}}
\vspace{0.45em}

{\normalsize\rmfamily\color{microsoftdark}
Weijie Wang$^{1,*}$ \hspace{0.7em}
Haoyu Zhao$^{1,*}$ \hspace{0.7em}
Yifan Yang$^{2}$ \hspace{0.7em}
Feng Chen$^{3}$ \hspace{0.7em}
Zeyu Zhang$^{1}$\\[-0.1em]
Yefei He$^{1}$ \hspace{0.7em}
Zicheng Duan$^{3}$ \hspace{0.7em}
Donny Y. Chen$^{4}$ \hspace{0.7em}
Yuqing Yang$^{2}$ \hspace{0.7em}
Bohan Zhuang$^{1}$\par
}
\vspace{0.2em}
{\footnotesize\rmfamily\color{microsoftgray}
$^{1}$ Zhejiang University \quad
$^{2}$ Microsoft Research \quad
$^{3}$ Adelaide University \quad
$^{4}$ Monash University\par
}
\end{center}
\blfootnote{$^*$ Equal contribution.}
\vspace{-0.05cm}

\begin{center}
    \centering
    \captionsetup{type=figure}
    \includegraphics[width=\linewidth]{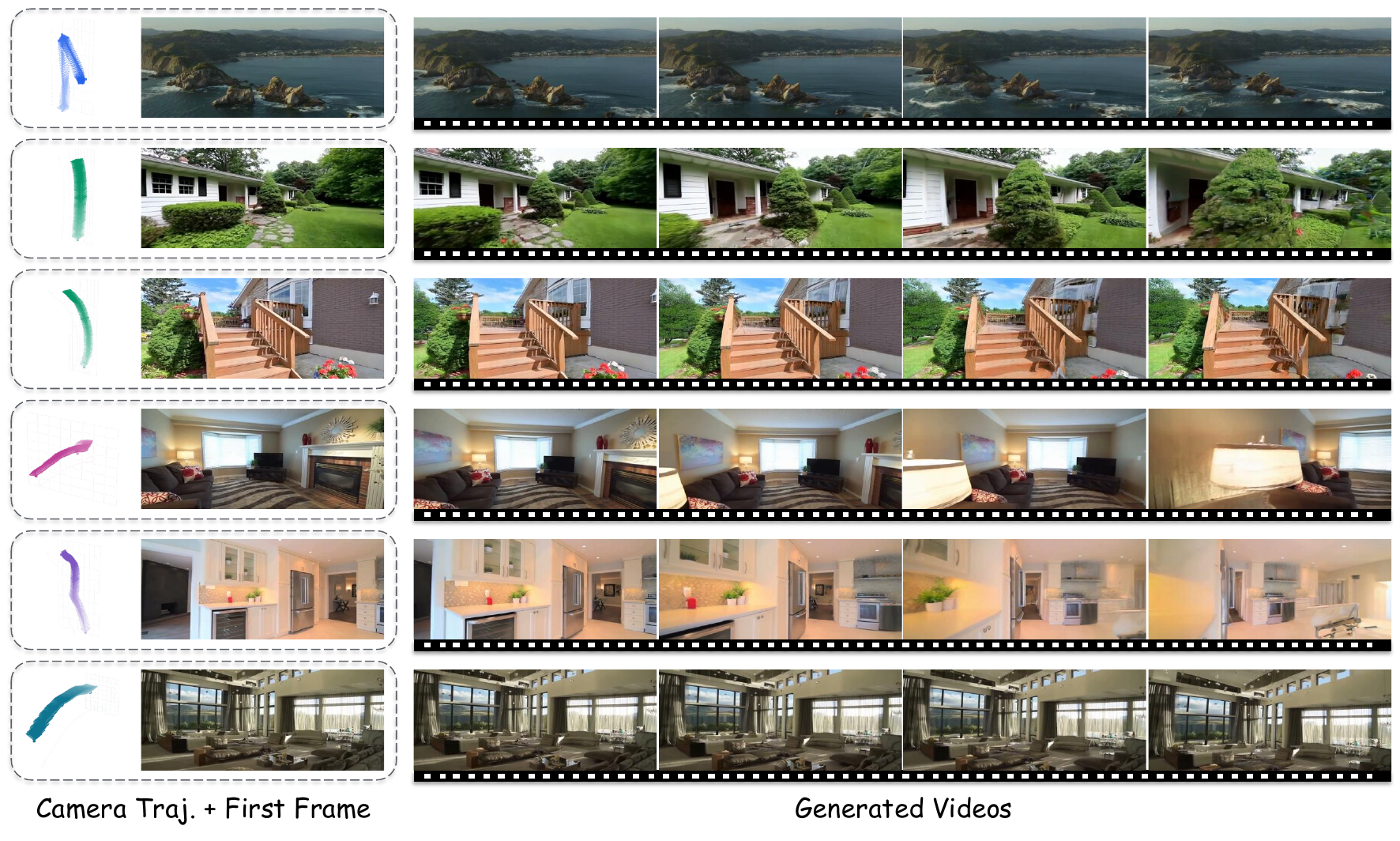}
    \vspace{-0.6cm}
    \captionof{figure}{\textbf{Geometrically consistent videos generated by \method{} with latent spatial memory.}
    Given a single input image and a user-specified camera trajectory (left), \method{} preserves spatial consistency by caching 3D information directly in the \emph{latent} space, rather than as an RGB-colored point cloud. This design enables memory queries through a single latent-resolution projection, avoiding the costly render-and-encode round trip required by prior RGB point cloud memories. Consequently, \method{} can faithfully return to previously observed regions even after large camera movements, while achieving up to $\mathbf{10.57\times}$ faster end-to-end video generation and $\mathbf{55\times}$ lower GPU memory usage than RGB-cache baselines.}
    \label{fig:teaser}
\end{center}

\vspace{-0.55em}
\begin{microsofttitlebox}
\setlength{\parindent}{0cm}
\setlength{\parskip}{0.015cm}
\raggedright
\nohyphens

\scriptsize
\setlength{\parindent}{0cm}
\setlength{\parskip}{0.01cm}
\begin{abstract}
Video world models that maintain 3D spatial consistency across generated frames typically rely on explicit point cloud memory constructed in RGB space. This design is both computationally expensive, requiring repeated rendering and VAE encoding, and inherently lossy, as the round trip through pixel space discards rich features of the learned latent representation. In this paper, we introduce \emph{latent spatial memory} for video world models, a persistent 3D cache that stores scene information directly in the diffusion latent space, avoiding pixel-space reconstruction. Building on this, we propose \method{}, a latent-space spatial memory framework that constructs the memory by lifting latent tokens into 3D via depth-guided back-projection and queries it by synthesizing novel views through direct latent-space warping.  This unified formulation eliminates both the information loss of pixel-space reconstruction and the computational burden of repeated encoding and rendering. Experiments show that latent spatial memory achieves up to \textbf{10.57}$\times$ faster end-to-end video generation and \textbf{55}$\times$ lower memory footprint relative to explicit 3D baselines. Leveraging the geometric prior of the diffusion model, \method{} attains state-of-the-art performance on WorldScore and strong reconstruction quality on RealEstate10K.
\end{abstract}

\vspace{0.08cm}
{\setlength{\parskip}{0.1cm}\scriptsize
{\microsoftmetalabel{Project Page}\href{https://aka.ms/latent-spatial-memory}{aka.ms/latent-spatial-memory}\par}
{\microsoftmetalabel{Keywords}\lsmkeywords\par}
{\microsoftmetalabel{Date}June 8, 2026\par}
}
\end{microsofttitlebox}

\clearpage
\section{Introduction}

Large-scale video diffusion models~\cite{sora,wan2025wan,kong2024hunyuanvideo,yang2024cogvideox,blattmann2023stable} have demonstrated remarkable ability to synthesize photorealistic sequences, motivating their use as world simulators that internalize visual dynamics and generate plausible future observations conditioned on camera trajectories or actions~\cite{bruce2024genie,valevski2024diffusion,alonso2024diffusion}. A central challenge in this paradigm is maintaining long-term scene consistency: without explicit spatial memory, even powerful generators accumulate geometric drift, producing frames that are individually convincing but collectively inconsistent when projected into a shared world coordinate system.
For example, when revisiting a previously generated region, the visual content is likely to have changed.

\begin{figure}[!t]
    \centering
    \includegraphics[width=\linewidth]{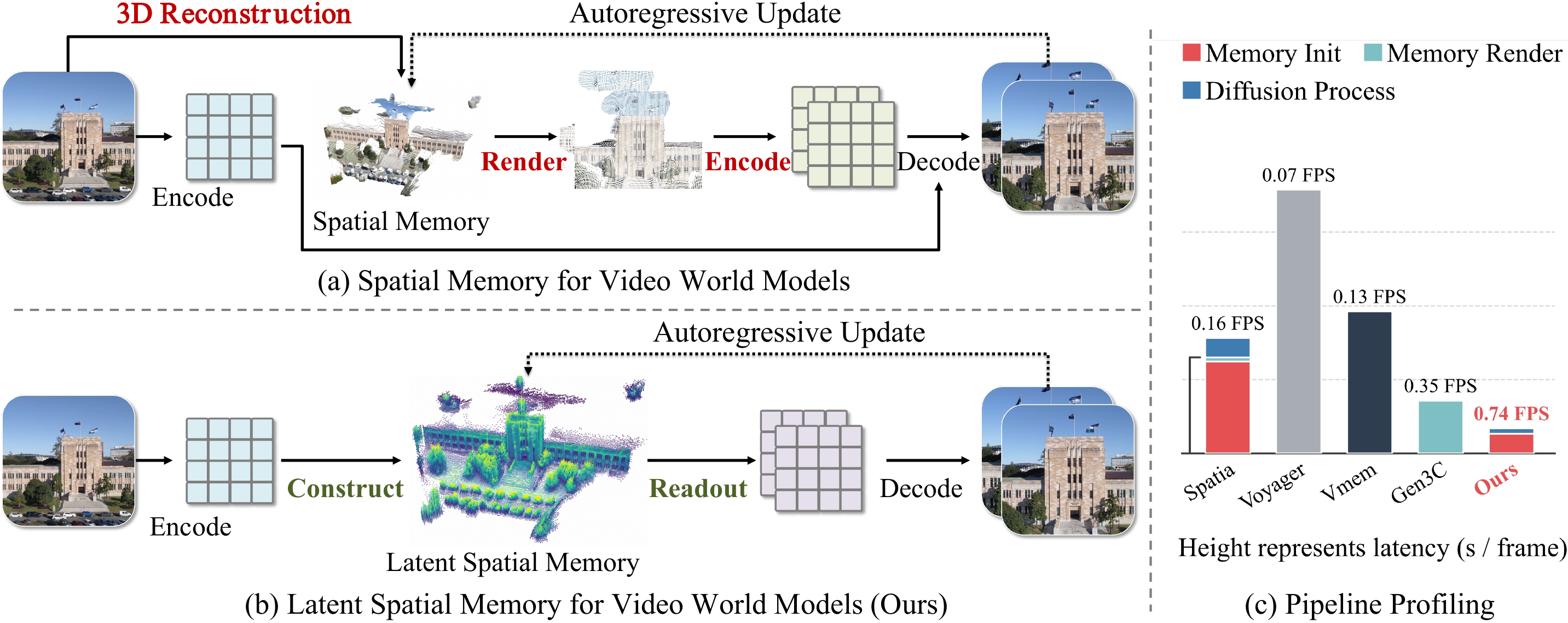}
    \caption{\textbf{Latent spatial memory vs.\ RGB point cloud based memory.}
(a): prior systems store memory in RGB point clouds and pay a render-and-encode round trip at every conditioning step.
(b): latent spatial memory is stored in diffusion latent space and can be read through simple projections, eliminating the per-step pixel-space detour and shrinking the cache footprint by the squared VAE compression factor.
(c): pipeline profiling shows that maintaining explicit memory is a major efficiency bottleneck.
}
    \label{fig:concept}
\end{figure}

A natural remedy is to equip the generator with a persistent 3D memory that tracks what has been observed. As shown in Fig~\ref{fig:concept}, recent world-generation systems~\cite{zhao2026spatia,huang2025voyager,yu2024wonderworld,yu2024wonderjourney,wang2025drivegen3d,wang2026world,duan2026liveworld,wu2025video,chou2025captain} adopt this strategy by maintaining an explicit point cloud in RGB space which is rendered and re-encoded at every step. While effective at enforcing scene consistency, this pipeline introduces two fundamental bottlenecks. As shown in Fig~\ref{fig:concept}(c), first, the repeated round trip between latent and pixel space is computationally prohibitive, as rendering millions of colored points at full resolution and re-encoding them dominates wall-clock time. Second, the RGB detour does not preserve the model's native latent conditioning features. RGB point cloud memories render a target-view image and then re-encode it into a latent tensor, which may be distorted by VAE reconstruction error and distribution mismatch.

To address these bottlenecks, we propose \textbf{latent spatial memory}, a persistent 3D cache that stores the diffusion model's latent features at world-space locations instead of RGB colors. As illustrated in Fig.~\ref{fig:concept}, an observed frame is first encoded into VAE latent space, and each latent-grid cell is lifted into 3D by depth-guided back-projection. Each memory element thus pairs a world-space coordinate with the corresponding frame latent token. At readout time, the cache is projected directly onto the target camera grid at latent resolution with depth-aware visibility handling, yielding a target-view latent tensor in the same space consumed by the diffusion backbone. This avoids both pixel-resolution rendering of the cache and per-step VAE encoding of the rendered image, eliminating the two main bottlenecks of RGB point cloud memories.

Building on this representation, we present \method{}, a video world model that generates long, geometrically consistent rollouts chunk by chunk through an initialize-readout-update latent memory cycle as shown in Fig~\ref{fig:pipeline}. 
First, \method{} initializes the latent spatial memory by encoding the initial frame and back-projecting its latent tokens into the world-space coordinates via depth-guided lifting~\cite{lin2025depth,wang2026feed,wang2025zpressor,wang2025volsplat,wang2026trisplat}.
For each subsequent chunk, it reads from memory by projecting the cache onto every target camera pose, producing latent-space feature tensors that are injected into the diffusion backbone through a ControlNet-style side branch~\cite{zhang2023adding}. 
After decoding the denoised latents into output frames, \method{} updates the memory by segmenting dynamic objects~\cite{carion2025sam3segmentconcepts}, estimating depth and back-projecting them into the cache to preserve geometric coherence. 
This cycle repeats over chunks to support long-horizon generation.

Extensive experiments validate the benefits of latent spatial memory in both generation quality and efficiency. Across WorldScore~\cite{worldscore} and RealEstate10K~\cite{zhou2018stereo}, \method{} achieves state-of-the-art or competitive performance against strong baselines. At the same time, end-to-end video generation is up to $\mathbf{10.57\times}$ faster and uses up to $\mathbf{55\times}$ less GPU memory in 3D cache than RGB point cloud readout, making world-consistent generation practical for long trajectories.
Our contributions can be summarized as follows:
\begin{itemize}
    \item We introduce \emph{latent spatial memory}, the 3D memory for video world models that operates entirely in latent space, avoiding heavy pixel-space conversion.
    \item We develop a data engine for collecting video clips with latent spatial memory. Building on this, we propose \method{}, a video world model featuring a ControlNet-style conditioning branch and a two-stage training strategy.
    \item Our method achieves state-of-the-art world generation on WorldScore and competitive novel view synthesis on RealEstate10K, with up to $\mathbf{10.57\times}$ end-to-end speedup and $\mathbf{55\times}$ lower memory usage.
\end{itemize}

\section{Related Work}

\subsection{Video Diffusion Models}
The diffusion and flow-matching frameworks~\cite{lipman2022flow,esser2024scaling} have rapidly advanced from high-fidelity image synthesis to temporally coherent video generation. Early video diffusion methods extended image architectures by interleaving temporal attention or 3D convolution layers to capture inter-frame dependencies~\cite{blattmann2023stable,guo2023animatediff,chen2024videocrafter2}, while later systems scaled this paradigm to photorealistic, multi-second outputs through large-scale pretraining on diverse video corpora~\cite{sora,kong2024hunyuanvideo,yang2024cogvideox,xu2024easyanimate,allegro2024,fan2025vchitect}. A common strategy is to operate within a compressed VAE latent space, which significantly reduces the computational cost of the denoising process and enables generation at higher resolutions and longer durations~\cite{blattmann2023stable,hacohen2024ltx,wan2025wan}. More recently, various autoregressive and streaming extensions have been proposed to generate longer sequences~\cite{chen2024diffusion,henschel2024streamingt2v,gu2025long,chen2025skyreels,xie2024progressive}. Despite these advances, most video diffusion models treat the generation process as fundamentally two-dimensional, in that frames are synthesized sequentially or in parallel without an explicit model of the underlying 3D scene geometry. As a consequence, generated videos frequently exhibit geometric drift, parallax violations, and inconsistent scene structure when subjected to large camera motions or extended temporal horizons.

\subsection{Camera-controllable Video Generation}
A parallel line of work seeks to condition video diffusion models on explicit camera trajectories to enable controllable viewpoint changes. Representative methods use additional control modules~\cite{he2024cameractrl}, attention mechanisms~\cite{yu2024viewcrafter,zhou2025stable}, or 3D-aware rendering signals fed as conditioning inputs~\cite{ren2025gen3c}. While these approaches offer fine-grained camera control within a single generation pass, they do not maintain a persistent scene representation across generation steps. Each clip is generated independently of previously synthesized content, so revisiting a region or extending a trajectory beyond the context window of the model may introduce inconsistencies. The absence of a shared spatial memory limits their applicability to long-horizon, exploratory world generation where 3D coherence must be preserved across many sequential generation rounds.

\subsection{Spatial Memory for Video Generation}
To maintain spatial consistency, a growing body of work augments video diffusion pipelines with explicit spatial memory structures that maintain a 3D scene representation across generation steps. A representative strategy is to lift observed or generated RGB frames into point clouds using estimated depth, accumulate them into a persistent 3D cache, and render target-view conditioning images at each step~\cite{yu2024wonderjourney,engstler2024invisible,yu2024wonderworld,huang2025voyager,zhao2026spatia,chen2025flexworld}. More recent systems further incorporate surfel-indexed view memory~\cite{li2025vmem} to improve temporal consistency over extended sequences. Such memory-based approaches have demonstrated clear improvements in multi-view coherence, since the 3D cache anchors each new frame to a shared geometric scaffold. Nevertheless, existing spatial memory designs operate entirely in RGB pixel space, which is both computation-intensive and is vulnerable to accumulated errors in repeated encoding-decoding operations. These limitations motivate our approach, which constructs and queries spatial memory entirely within the latent space of the diffusion model, preserving representational fidelity while eliminating the rendering and re-encoding overhead that constrains prior methods.

\section{Preliminaries}
\label{sec:prelim}

A video world model synthesizes a multi-view-consistent sequence $\{I^t\}_{t=1}^{T}$ from an initial frame $I^0$ and a camera trajectory $\{(\mathbf{E}^t, K^t)\}$, where $\mathbf{E}^t$ and $K^t$ are the extrinsics and intrinsics of frame $t$.
Modern systems build on a pretrained video diffusion backbone that operates in the latent space of a VAE with encoder $\mathcal{E}$, decoder $\mathcal{D}$, spatial stride $s$, and latent channel count $C$.
Generation proceeds autoregressively over the overlapping chunk of latent frames, each denoised from Gaussian noise conditioned on its predecessors.
Although effective for short clips, this autoregressive scheme conditions only on a small temporal context, so that information about earlier observations gradually fades as the rollout advances.
As a consequence, geometric drift accumulates whenever the camera revisits a previously observed region: the same physical surface may reappear at a different position, with inconsistent texture, or with altered scene layout, breaking the multi-view consistency that downstream applications rely on~\cite{yang2024cogvideox, wan2025wan}.

\begin{figure*}[t]
    \centering
    \includegraphics[width=\linewidth]{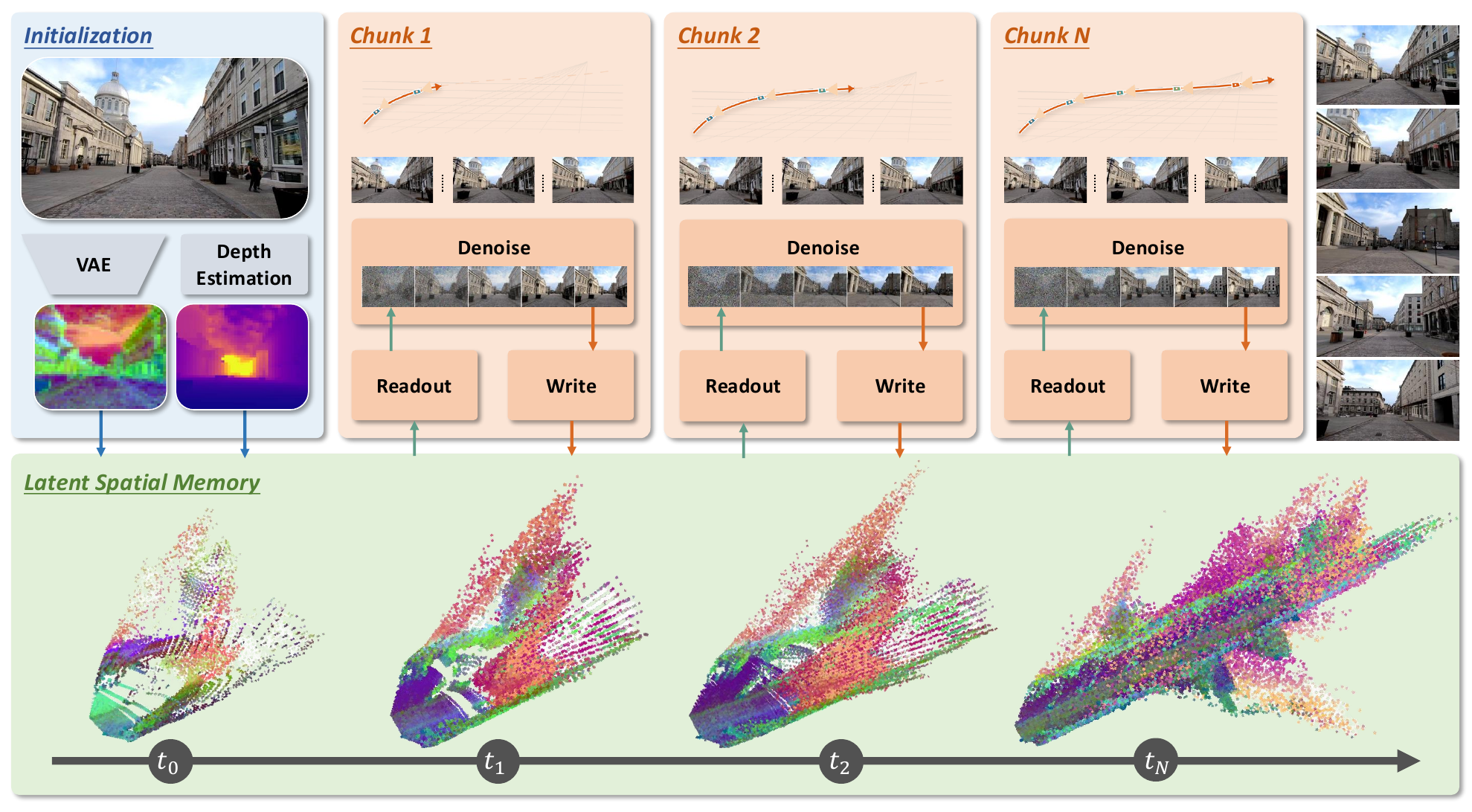}
    \caption{\textbf{Overview of \method{}.} \method{} initializes a 3D latent memory from $I_0$ by encoding it into VAE latents and lifting them with depth-guided back-projection. At each target view, the memory is read through a latent-resolution projection, and generation proceeds chunk by chunk: each decoded output chunk is re-estimated for depth, re-encoded into latents, and back-projected to extend the memory.}
    \label{fig:pipeline}
\end{figure*}

A common remedy is to attach a persistent \emph{RGB point cloud}~\cite{zhao2026spatia, huang2025voyager, yu2024wonderworld}
\begin{equation}
\mathcal{M}_{\text{rgb}} = \bigl\{(\mathbf{p}_i, \mathbf{c}_i)\bigr\}, \qquad \mathbf{p}_i \in \mathbb{R}^3,\; \mathbf{c}_i \in [0, 1]^3,
\label{eq:rgb-mem}
\end{equation}
that records the color $\mathbf{c}_i$ of every observed surface point $\mathbf{p}_i$.
The memory is initialized from $I^0$ through depth-guided back-projection and grown along the rollout, providing a long-term geometric scaffold that the autoregressive context alone cannot maintain.
Conditioning on this 3D memory, however, forces a full pixel-space round trip at every target view: the memory is first rasterised into an RGB image at the target pose $(\mathbf{E}^t, K^t)$, and then re-encoded into a latent tensor $\hat{\mathbf{z}}^t$ that is supplied to the denoiser,
\begin{equation}
\hat{\mathbf{z}}^t = \mathcal{E}\,\!\bigl (\mathrm{Rasterise}(\mathcal{M}_{\text{rgb}};\, \mathbf{E}^t, K^t)\bigr),
\label{eq:rgb-read}
\end{equation}
where $\mathrm{Rasterise}(\cdot)$ denotes $z$-buffered projection followed by shading.
This conditioning step introduces two major costs. First, the rasterizer and VAE encoder operate at pixel resolution, while the generator consumes latent-resolution tensors, making each read unnecessarily expensive and increasingly costly as the memory grows. Second, RGB readout must be re-encoded into a latent signal, which can deviate from the model's native latent representation due to reconstruction error and distribution mismatch. Thus, storing memory in pixel space creates both computational and representation bottlenecks for a latent-space generator.

\section{Method}
\label{sec:method}

\subsection{Overview}
\label{sec:method:principle}

\method{} maintains a persistent latent memory $\mathcal{M}$ and generates videos by first initializing memory and then repeating a readout-update cycle over chunks,
as illustrated in Figure~\ref{fig:pipeline}.
\textbf{Initialization:} the initial frame $I^0$ is encoded by encoder $\mathcal{E}$ and lifted
into 3D space via depth-guided back-projection, seeding $\mathcal{M}$
with one latent-attributed 3D point per latent cell
(Section~\ref{sec:method:mem}).
\textbf{Readout and denoising:}
For each chunk, $\mathcal{M}$ is projected onto the target camera grids at latent resolution, producing target-view latent feature tensors. 
These conditional tensors are injected into the diffusion backbone through a ControlNet-style side branch, allowing the spatial memory to work entirely in latent space
(Section~\ref{sec:method:read}).
\textbf{Memory update:} the generated frames are re-encoded and back-projected to update
$\mathcal{M}$, with moving objects and sky excluded to preserve geometric
coherence (Section~\ref{sec:method:update}).
Crucially, $\mathcal{M}$ is queried directly in latent space, while
pixel-space operations appear only during chunk-level memory update. This
avoids the redundant render-and-encode loop of RGB point cloud
memories and reduces both readout cost and memory footprint by a factor of
$s^2$.

\subsection{Latent Spatial Memory Initialization}
\label{sec:method:mem}

We represent the memory as a set of latent-attributed 3D points
\begin{equation}
\mathcal{M} = \bigl\{(\mathbf{p}_i, \mathbf{f}_i)\bigr\},
\qquad \mathbf{p}_i \in \mathbb{R}^{3},\; \mathbf{f}_i \in \mathbb{R}^{C},
\label{eq:lat-mem}
\end{equation}
where each point $i$ pairs a world-space coordinate $\mathbf{p}_i$ with
latent feature $\mathbf{f}_i$ drawn directly from the VAE encoder, matching the native
input space of the diffusion backbone.

\paragraph{Construction.}
Given a latent frame $\mathbf{z} \in \mathbb{R}^{C \times h \times w}$,
metric depth map $D$, intrinsics $K$, and camera pose $\mathbf{E}$
from a feed-forward reconstructor~\cite{lin2025depth}, we first
downsample $D$ to the latent grid and rescale $K$ accordingly; in what
follows we use $D$ and $K$ to refer to their latent-resolution versions.
Each latent cell $(u,v)$ is then back-projected into world space,
producing one memory element per cell:
\begin{equation}
\mathbf{p}_{uv} = \pi^{-1}(u, v, D(u,v);\, K, \mathbf{E}), \qquad \mathbf{F}_{uv} = \mathbf{z}[:, v, u],
\label{eq:lift}
\end{equation}
where $\pi^{-1}$ denotes standard pinhole back-projection
(see the accompanying appendix for details) and $\mathbf{F}_{uv}\in \mathbb{R}^C$ denotes the latent token stored at the memory element anchored at $\mathbf{p}_{uv}$.
The initial memory is seeded by applying Eq.~\ref{eq:lift} to the
encoded latent of $I^0$, and is subsequently updated by the following autoregressive memory update.

\subsection{Latent Spatial Memory Readout}
\label{sec:method:read}

 Given a target view $(\mathbf{E}^t, K^t)$, we query the latent memory
$\mathcal{M}$ by projecting all memory points onto the target camera grid
at latent resolution. For each target cell, we retain the frontmost
projected point using $z$-buffering and retrieve its associated latent
token:
\begin{equation}
i^t(u,v)
=
\operatorname*{arg\,min}_{i \in \Omega^t(u,v)}
\bigl[\mathbf{E}^t \mathbf{p}_i\bigr]_z,
\qquad
\hat{\mathbf{z}}^t(u,v)
=
\mathbf{F}_{i^t(u,v)} .
\label{eq:read}
\end{equation}
Here $\Omega^t(u,v)$ denotes the set of memory points that project to
latent cell $(u,v)$ with positive depth under the target view
$(\mathbf{E}^t,K^t)$, and $[\cdot]_z$ extracts the depth coordinate.
Cells with $\Omega^t(u,v)=\emptyset$ are zero-filled. We also produce a
binary visibility mask $\mathbf{m}^t\in\{0,1\}^{h\times w}$ indicating
which cells receive at least one projected point. This mask allows the
denoiser to distinguish genuinely unseen regions from observed regions
whose latent feature is zero. 

This readout preserves the stored conditioning signal: when the memory is
constructed from a source frame and queried from the same view,
Eq.~\ref{eq:read} retrieves the corresponding source-view latent tokens on
visible cells, up to discretization and occlusion. 
To generate each chunk, the latent memory readouts $\hat{\mathbf{z}}^t$ and visibility masks $\mathbf{m}^t$ are concatenated and passed to a ControlNet-style side branch~\cite{zhang2023adding}, which injects memory features into the video diffusion backbone. 
Besides, segment-aware rotary encodings are also leveraged to mark noisy target, clean preceding, and clean reference frames in a single forward pass.
Because the readouts already lie in the backbone's latent space, no redundant render-encode loops are needed, which are required by pipelines based on RGB point cloud memory.

\subsection{Autoregressive Memory Update}
\label{sec:method:update}

After each chunk is generated, \method{} updates $\mathcal{M}$ with the
newly observed scene content. 
We estimate depth and camera parameters for the generated frames using a
feed-forward reconstructor~\cite{lin2025depth} and
back-project the corresponding latent tokens into the memory using Eq. \ref{eq:lift}. Then, the updated memory can be denoted as:
\begin{equation}
\mathcal{M}
\leftarrow
\mathcal{M}
\cup
\bigl\{(\mathbf{p}_{uv}, \mathbf{F}_{uv})\bigr\}_{(u,v)\in\Lambda^t},
\label{eq:lat-update}
\end{equation}
where $\Lambda^t$ contains latent cells with valid depth outside dynamic
objects and sky regions, detected by an open-vocabulary entity
extractor~\cite{yang2025qwen3} and a video segmenter~\cite{carion2025sam3segmentconcepts}.
This filtering prevents transient or geometrically unreliable content
from contaminating the persistent memory. 
The current chunk latents are also carried forward as short-term temporal
context for the next chunk.

\subsection{Efficient Adaptation to Existing Diffusion Models}
We adapt a pretrained camera-controllable video diffusion
transformer~\cite{wan2025wan,zhao2026spatia} in two stages. 
In the first stage, we freeze the backbone and VAE and train only the
ControlNet-style side branch, aligning latent memory readouts with the
backbone feature space without perturbing the pretrained generative
prior. 
In the second stage, we freeze the side branch and attach rank-32 LoRA
adapters~\cite{hu2021lora} ($\alpha=32$) to the backbone. We optimize only these adapters, enabling lightweight adaptation to the memory condition while
preserving the backbone's appearance and motion priors. 
Both stages use the flow-matching objective~\cite{lipman2022flow} on the
target frames.

\section{Experiments}
\label{sec:exp}

\begin{table*}
    \centering
    \small
    \setlength{\tabcolsep}{1.5pt}
    \caption{
    \textbf{Evaluation Results on WorldScore~\cite{worldscore}.}
    The \textbf{Average Score} column is the average of the Static Score and Dynamic Score, while the remaining metrics are computed by the WorldScore benchmark.
    }
    \resizebox{\textwidth}{!}{%
        \begin{tabular}{>{\raggedright\arraybackslash}p{88pt}|ccc|ccc|cccc|ccc}
        \toprule
        Method
        & \makecell{\textbf{Average}\\\textbf{Score}}
        & \makecell{\textbf{Static}\\\textbf{Score}}
        & \makecell{\textbf{Dynamic}\\\textbf{Score}}
        & \makecell{Camera\\Ctrl}
        & \makecell{Object\\Ctrl}
        & \makecell{Content\\Align}
        & \makecell{3D\\Const}
        & \makecell{Photo\\Const}
        & \makecell{Style\\Const}
        & \makecell{Subject\\Quality}
        & \makecell{Motion\\Acc}
        & \makecell{Motion\\Mag}
        & \makecell{Motion\\Smooth} \\
        \midrule
        \multicolumn{14}{l}{\textit{Models with 3D Cache}} \\
        WonderJourney     & 54.19 & 63.75 & 44.63 & 84.60 & 37.10 & 35.54 & 80.60 & 79.03 & 62.82 & \underline{66.56} & - & - & - \\
        InvisibleStitch   & 51.95 & 61.12 & 42.78 & \textbf{93.20} & 36.51 & 29.53 & 88.51 & 89.19 & 32.37 & 58.50 & - & - & - \\
        WonderWorld       & 61.79 & 72.69 & 50.88 & \underline{92.98} & 51.76 & 71.25 & 86.87 & 85.56 & 70.57 & 49.81 & - & - & - \\
        Voyager           & 66.08 & \textbf{77.62} & 54.53 & 85.95 & \underline{66.92} & 68.92 & 81.56 & 85.99 & \underline{84.89} & \textbf{71.09} & - & - & - \\
        FlashWorld        & 60.23 & 70.85 & 49.60 & 84.43 & 50.28 & 56.54 & 85.87 & 86.72 & 79.36 & 52.75 & - & - & - \\
        LucidDreamer      & 59.84 & 70.40 & 49.28 & 88.93 & 41.18 & \textbf{75.00} & \underline{90.37} & \underline{90.20} & 48.10 & 58.99 & - & - & - \\
        Spatia            & \underline{69.73} & 72.63 & \underline{66.82} & 75.66 & 52.32 & 69.95 & 86.40 & 89.10 & 80.09 & 54.86 & 54.83 & 24.75 & \underline{80.26} \\
        \midrule
        \multicolumn{14}{l}{\textit{General Video Models}} \\
        VideoCrafter2     & 50.03 & 52.57 & 47.49 & 28.92 & 39.07 & \underline{72.46} & 65.14 & 61.85 & 43.79 & 56.74 & 47.12 & 30.40 & 29.39 \\
        EasyAnimate       & 52.25 & 52.85 & 51.65 & 26.72 & 54.50 & 50.76 & 67.29 & 47.35 & 73.05 & 50.31 & \underline{75.00} & 31.16 & 40.32 \\
        Allegro           & 53.64 & 55.31 & 51.97 & 24.84 & 57.47 & 51.48 & 70.50 & 69.89 & 65.60 & 47.41 & 54.39 & \textbf{40.28} & 37.81 \\
        CogVideoX-I2V     & 60.64 & 62.15 & 59.12 & 38.27 & 40.07 & 36.73 & 86.21 & 88.12 & 83.22 & 62.44 & 69.56 & 26.42 & 60.15 \\
        Vchitect-2.0      & 40.38 & 42.28 & 38.47 & 26.55 & 49.54 & 65.75 & 41.53 & 42.30 & 25.69 & 44.58 & 33.59 & \underline{33.81} & 21.31 \\
        LTX-Video         & 55.99 & 55.44 & 56.54 & 25.06 & 53.41 & 39.73 & 78.41 & 88.92 & 53.50 & 49.08 & \textbf{76.22} & 29.95 & 71.09 \\
        Wan2.1            & 55.21 & 57.56 & 52.85 & 23.53 & 40.32 & 45.44 & 78.74 & 78.36 & 77.18 & 59.38 & 54.27 & 33.26 & 38.05 \\
        \midrule
        \method{} (Ours)                             & \textbf{70.36} & \underline{73.60} & \textbf{67.11} & 55.36 & \textbf{74.17} & 42.09 & \textbf{92.21} & \textbf{93.95} & \textbf{96.91} & 60.50 & 51.36 & 24.18 & \textbf{80.36} \\
        \bottomrule
        \end{tabular}
    }
    \label{tab:world-score}
\end{table*}

\subsection{Experimental Setup}
\label{sec:exp:setup}

\paragraph{Datasets and baselines.}
We train on a corpus of videos from RealEstate10K~\cite{zhou2018stereo}, with depth and camera poses annotated by~\cite{huang2025vipe,lin2025depth} and dynamic regions removed as in Sec.~\ref{sec:method:update}. Evaluation is performed on WorldScore~\cite{worldscore}, which reports ten metrics covering controllability, consistency, quality, and motion, and on RealEstate10K, which provides paired ground truth for novel view synthesis and supports the closed loop protocol of Spatia~\cite{zhao2026spatia}. On WorldScore we compare against RGB point cloud based scene generators~\cite{yu2024wonderjourney,engstler2024invisible,yu2024wonderworld,huang2025voyager} and state of the art foundation video generators~\cite{chen2024videocrafter2,xu2024easyanimate,allegro2024,yang2024cogvideox,fan2025vchitect,hacohen2024ltx,wan2025wan,zhao2026spatia}. On RealEstate10K we additionally compare against the view memory baselines SEVA~\cite{zhou2025stable} and VMem~\cite{li2025vmem}, and against 3D aware video generators ViewCrafter~\cite{yu2024viewcrafter} and FlexWorld~\cite{chen2025flexworld}.

\paragraph{Implementation and evaluation.}
Our backbone is Wan2.2-TI2V-5B~\cite{wan2025wan}, whose VAE has a compression ratio of $4\times16\times16$ and latent channel count $C=48$. Each generation chunk contains nine latent frames at resolution $44\times80$, corresponding to 33 RGB frames at $704\times1280$.
Training was conducted on 32 A100 GPUs with a global batch size of 64. The ControlNet-style side branch is initialized from its corresponding blocks of Wan2.2 and trained with an AdamW optimizer, bfloat16 mixed precision, and the flow-matching objective~\cite{lipman2022flow} on target frames. At inference, we use the UniPC flow scheduler with 40 sampling steps. We evaluate generation quality using the WorldScore Average Score and its constituent metrics, as well as PSNR, SSIM, LPIPS, and the closed-loop metrics PSNR$_C$, SSIM$_C$, and LPIPS$_C$ on RealEstate10K following Spatia. Efficiency is measured on a single NVIDIA H100 with the wall-clock time and peak GPU memory for one cache read as rollout length increases.

\begin{figure*}[!t]
    \centering
    \includegraphics[width=1.0\textwidth]{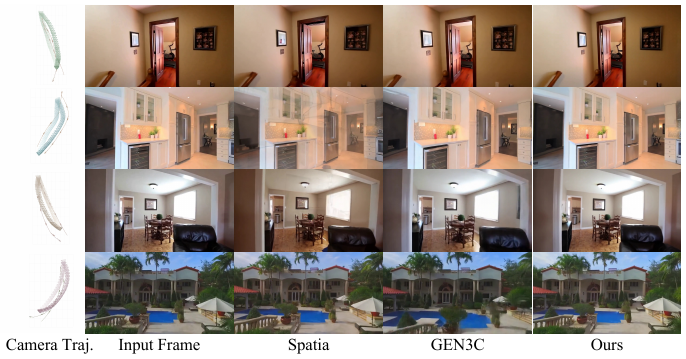}
    \caption{\textbf{Closed-loop revisit comparison on RealEstate10K.}
    In the closed-loop test, the camera trajectory gradually returns to its
    starting point. The comparison between the last frame and the input frame
    shows that \method{} maintains strong consistency under the revisit setting.}
    \label{fig:re10k_revisit}
\end{figure*}

\begin{table*}[!t]
    \centering
    \small
    \caption{
    \textbf{Evaluation Results on RealEstate10K~\cite{zhou2018stereo}.} We report novel-view synthesis and closed-loop results. Following Spatia~\cite{zhao2026spatia}, closed-loop evaluation measures the similarity between the initial frame and the final frame after generating a return trajectory.
    }
    \begin{tabular*}{\textwidth}{@{\extracolsep{\fill}}lccc|ccc@{}}
    \toprule
    Method & \multicolumn{3}{c|}{Novel View Synthesis} & \multicolumn{3}{c}{Closed-Loop} \\
    \cmidrule(lr){2-4} \cmidrule(lr){5-7}
    & PSNR $\uparrow$ & SSIM $\uparrow$ & LPIPS $\downarrow$ & PSNR$_{C}$ $\uparrow$ & SSIM$_{C}$ $\uparrow$ & LPIPS$_{C}$ $\downarrow$ \\
    \midrule
    SEVA \cite{zhou2025stable}           & 13.07 & 0.515 & 0.445 & - & - & - \\
    VMem \cite{li2025vmem}               & 14.62 & 0.522 & 0.426 & - & - & - \\
    \midrule
    ViewCrafter \cite{yu2024viewcrafter} & 15.78 & 0.580 & 0.396 & 14.79 & 0.481 & 0.365 \\
    FlexWorld \cite{chen2025flexworld}   & 16.25 & 0.593 & 0.370 & 12.20 & 0.428 & 0.598 \\
    Voyager \cite{huang2025voyager}      & 17.79 & 0.636 & 0.297 & 17.66 & 0.540 & 0.380 \\
    Spatia \cite{zhao2026spatia}         & \textbf{18.58} & \underline{0.646} & \underline{0.254} & \underline{19.38} & \underline{0.579} & \textbf{0.213} \\
    \midrule
    \method{} (Ours)                     & \underline{18.38} & \textbf{0.779} & \textbf{0.250} & \textbf{20.05} & \textbf{0.825} & \underline{0.228} \\
    \bottomrule
    \end{tabular*}
    \label{tab:re10k-merge}
\end{table*}

\subsection{Main Results}
\label{sec:exp:main}

\begin{figure*}[!t]
    \centering
    \includegraphics[width=1.0\textwidth]{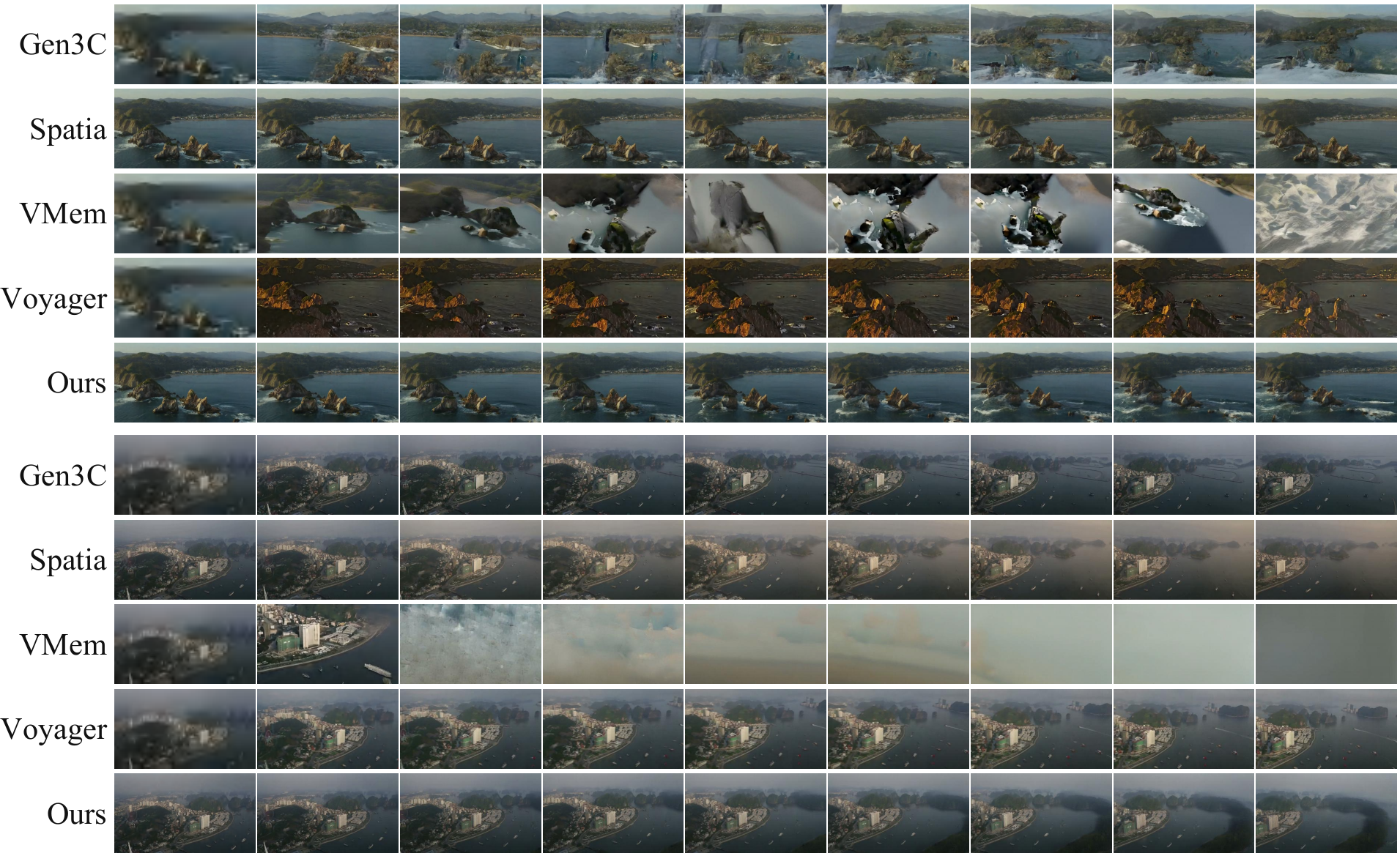}
    \caption{\textbf{Open-domain video comparison.}
    Generations on out-of-domain outdoor and natural scenes that differ substantially from the training distribution. \method{} generalizes effectively, producing temporally smooth and 3D-consistent rollouts under aggressive camera motion, whereas RGB point-cloud baselines exhibit stretched textures on unseen layouts and severe visual artifacts.}
    \label{fig:open_domain_video}
\end{figure*}

\paragraph{World generation on WorldScore.}
Table~\ref{tab:world-score} summarises results on WorldScore. \method{} attains the highest Average Score of all compared systems, improving over the memory augmented Spatia baseline and clearly outperforming every foundation video generator that lacks a persistent spatial representation. The advantage is strongest on the dynamic partition, while \method{} remains competitive on the static partition. On the static axis, \method{} leads on 3D and photometric consistency, confirming that latent spatial memory preserves the geometric scaffolding that RGB caches provide without incurring their representational loss. We attribute this balance to two design choices. First, the readout in Eq.~\ref{eq:read} injects geometric hints at the same resolution and distribution as the native latents of the backbone, so the generator does not have to reconcile two incompatible signal spaces. Second, the dynamic object filter described in Section~\ref{sec:method:update} prevents moving elements from contaminating the persistent memory, which is a common source of drift in RGB point cloud pipelines. Figure~\ref{fig:open_domain_video} shows side-by-side generations on out-of-domain prompts, where \method{} maintains 3D coherence on scenes that lie far from the RealEstate10K training distribution while RGB-cache and memory-free baselines exhibit visible drift.

\paragraph{Novel view synthesis and closed-loop consistency.}
Table~\ref{tab:re10k-merge} reports results on RealEstate10K. In the standard novel view synthesis setting, \method{} achieves the best SSIM and LPIPS, while remaining close to Spatia on PSNR, surpassing both view memory baselines and all 3D aware video generators. The closed loop setting is particularly informative because it amplifies long horizon drift, since any per step inaccuracy accumulates over a trajectory that returns to its starting viewpoint. Figure~\ref{fig:re10k_revisit} visualizes this return trajectory and the resulting agreement between the initial and revisited frames. \method{} achieves the best PSNR$_C$ and SSIM$_C$ under this protocol, while remaining second-best on LPIPS$_C$, indicating that latent spatial memory anchors the generator to a coherent geometric representation even after the camera has left and returned to a region. The supplementary appendix provides additional side-by-side video comparisons. These results show that moving the cache into latent space not only matches a well tuned RGB cache but improves on it, because the stored feature vectors carry semantic and textural information that three color channels cannot express.

\begin{figure*}
    \centering
    \includegraphics[width=\linewidth]{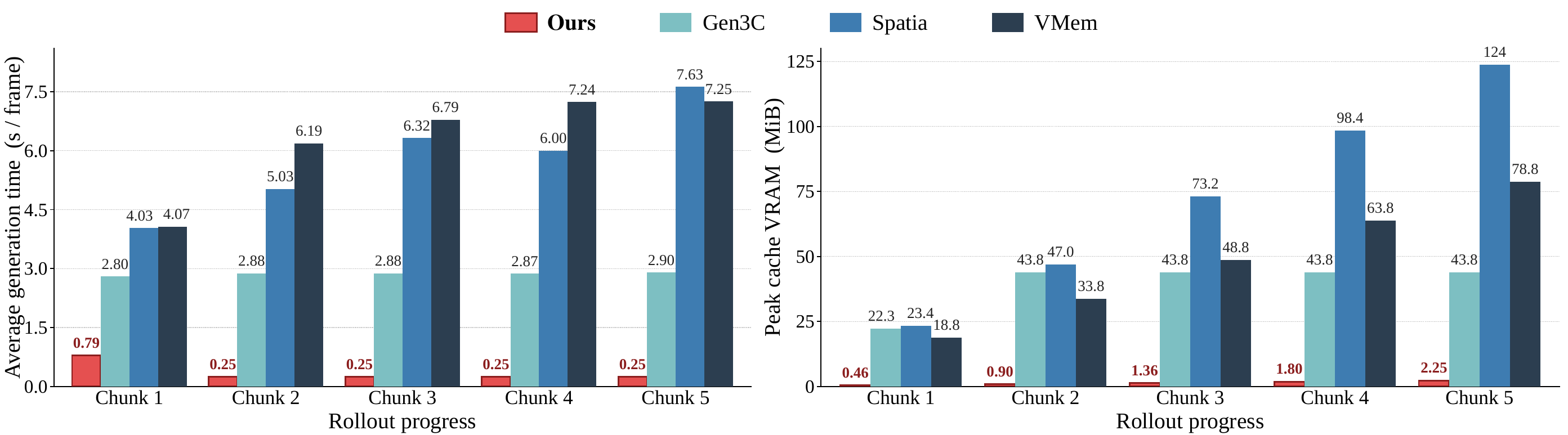}
    \caption{\textbf{Efficiency scaling with rollout progress.}
    Average cache-read time (\emph{left}) and peak cache footprint (\emph{right}) measured on a single NVIDIA H100 across five rollout chunks. After the first chunk amortises a one-off setup pass, \method{} settles at a per-frame cost of $0.25$\,s and a cache footprint that grows by less than $0.5$\,MiB per chunk. RGB-cache baselines require an order of magnitude more memory and one-to-two orders of magnitude more time per frame, since every conditioning step re-rasterises the accumulated point cloud and re-encodes the result through the VAE. Latent spatial memory removes this redundant trip, leaving the conditioning loop with a single latent-resolution projection.}
    \label{fig:efficiency_scaling}
\end{figure*}

\paragraph{Efficiency.}
A central motivation for latent spatial memory is to remove the pixel space round trip from the per step critical path. Figure~\ref{fig:efficiency_scaling} plots, as a function of rollout length, the wall clock time of a single cache read and the peak GPU memory required to maintain the cache. We compare \method{} with representative RGB point cloud baselines that render the accumulated cloud with a z-buffered renderer and re-encodes the resulting image through the VAE at every step. Both quantities grow slowly for latent spatial memory, because the read operates at the native latent resolution and the cache itself is smaller by a factor of $s^2$ per spatial dimension. In contrast, the RGB cache exhibits rapid growth along both axes, since the per step rendering scales with $HW$ rather than $hw$ and the cloud expands roughly linearly with the number of anchored frames. At matched rollout length, end-to-end video generation is $10.57\times$ faster than the RGB pipeline and consumes $55\times$ less GPU memory on 3D cache. The gap widens with horizon, and RGB baselines eventually exhaust the memory budget on trajectories that latent spatial memory completes comfortably.

\subsection{Ablation Studies}
\label{sec:exp:ablation}

We isolate the contribution of each component of \method{} on a split of WorldScore, so that ablations are directly comparable to the main benchmark. Component level results are reported in Table~\ref{tab:ablation}, and the sensitivity to the depth source used to build the latent cache is reported separately in Table~\ref{tab:depth}.

\paragraph{Latent vs. RGB memory.}
Replacing the latent cache with an RGB point cloud of the same source frames, while keeping the backbone and training recipe unchanged, decreases the Average Score and weakens 3D and photometric consistency. This confirms that bottleneck of the RGB detour discards information that backbone can exploit when the cache remains in latent space.

\paragraph{Feature downsampling vs. geometry downsampling.}
An alternative to the design in Eq.~\ref{eq:lift} is to upsample the latent feature to pixel resolution and lift at full resolution, then aggregate into the latent grid during readout. 
This variant degrades 3D consistency because it interpolates features that lie outside the backbone's pretraining distribution. 
Matching geometry to the native latent grid instead is both cheaper and more faithful to the generator.

\paragraph{Dynamic object filtering.}
Disabling the mask that excludes moving objects and the sky from cache updates degrades long horizon stability, since stale dynamic content persists in the memory and is splatted back into future chunks. The effect is largest on 3D and photometric consistency.

\paragraph{Two stage training.}
Replacing the two stage schedule with a single stage that jointly trains the side branch and LoRA reduces final quality, because the backbone adapts to an immature conditioning signal early in optimisation. Freezing the backbone during stage one and only then unlocking LoRA stabilises convergence.
\begin{table}
    \centering
    \small
    \setlength{\tabcolsep}{2pt}
    \caption{\textbf{Ablation studies on a WorldScore split.} Each row disables or alters a single component of \method{} while all other settings remain fixed.}
    \begin{tabular}{>{\raggedright\arraybackslash}p{0.38\linewidth}ccccc}
    \toprule
    Variant & \makecell{Avg\\$\uparrow$} & \makecell{Static\\$\uparrow$} & \makecell{Dynamic\\$\uparrow$} & \makecell{3D\\Cons $\uparrow$} & \makecell{Photo\\Cons $\uparrow$} \\
    \midrule
    \method{} (full)                        & \textbf{70.36} & \textbf{73.60} & \textbf{67.11} & \textbf{92.21} & \textbf{93.95} \\
    \midrule
    Explicit RGB Point Cloud                & \underline{67.71} & \underline{70.49} & \underline{64.93} & \underline{90.75} & \underline{91.10} \\
    Feature Upsample, Pixel Resolution Lift & 60.85 & 62.41 & 59.28 & 84.90 & 79.81 \\
    No Dynamic Object Filter                & 61.20 & 62.69 & 59.70 & 80.88 & 76.10 \\
    Single Stage Training                   & 63.18 & 65.15 & 61.20 & 87.11 & 84.47 \\
    \bottomrule
    \end{tabular}
    \label{tab:ablation}
\end{table}

\paragraph{Depth source.}
Because latent spatial memory is constructed from estimated depth, we study its robustness by swapping the default DepthAnything 3~\cite{lin2025depth} reconstructor for alternative depth predictors~\cite{keetha2025mapanything,piccinelli2024unidepth}. Table~\ref{tab:depth} shows that \method{} remains competitive under noisier depth, because the ControlNet style side branch treats the projected cache as a soft geometric hint rather than a hard constraint, and the dynamic filter removes the worst outliers before they enter the memory. The margin relative to the strongest depth estimator is modest, indicating that the benefits of latent spatial memory do not hinge on a particular reconstructor.
\begin{table}
    \centering
    \small
    \setlength{\tabcolsep}{2pt}
    \caption{\textbf{Sensitivity to the depth source.} We swap DepthAnything 3 for alternative depth predictors while keeping all other components of \method{} fixed.}
    \begin{tabular}{>{\raggedright\arraybackslash}p{0.38\linewidth}ccccc}
    \toprule
    Depth Source & \makecell{Avg\\$\uparrow$} & \makecell{Static\\$\uparrow$} & \makecell{Dynamic\\$\uparrow$} & \makecell{3D\\Cons $\uparrow$} & \makecell{Photo\\Cons $\uparrow$} \\
    \midrule
    DepthAnything 3~\cite{lin2025depth} (default) & 70.36 & 73.60 & 67.11 & 92.21 & 93.95 \\
    MapAnything~\cite{keetha2025mapanything}      & 69.66 & 72.78 & 66.53 & 91.89 & 93.32 \\
    UniDepth~\cite{piccinelli2024unidepth}        & 69.13 & 72.15 & 66.10 & 91.63 & 92.79 \\
    \bottomrule
    \end{tabular}
    \label{tab:depth}
\end{table}

\section{Conclusion}
\label{sec:conclusion}

We have introduced \emph{latent spatial memory}, a 3D cache that stores the video diffusion model's own latent features rather than RGB colors at each world-space point, and built \method{} around this representation as a video world model that operates entirely within the VAE latent manifold.
By reading the cache through a single latent-resolution projection, \method{} removes the render-and-encode round trip that dominates the per-step cost of RGB point cloud caches.
The decode-and-re-encode pair required to grow the cache is amortised over an entire chunk and never appears in the conditioning loop, so the per-step critical path is freed of pixel-space operations and the cache footprint shrinks by the squared VAE compression factor.
Across WorldScore and RealEstate10K, \method{} delivers state-of-the-art quality while reading from the cache an order of magnitude faster and with an order of magnitude less GPU memory than RGB-cache baselines, turning world-consistent generation into a process that scales with the rollout horizon.

\paragraph{Limitations and future work.}
The dynamic-region filter excludes moving entities from the persistent memory because their geometry is unreliable, so \method{} does not maintain the state of dynamic actors across chunks.
Scenes dominated by pervasive motion therefore benefit less from the cache than scenes dominated by rigid geometry, and persisting dynamic content across chunks is a natural direction for future work.
\vspace{1cm}

\clearpage
\bibliographystyle{unsrtnat}
\bibliography{reference}

\clearpage
\appendix

\section{Geometric Details}
\label{sec:app_geometry}

\paragraph{Conventions.}
Extrinsics $\mathbf{E}_t \in \mathrm{SE}(3)$ map world points into camera $t$, and we denote the camera-to-world inverse by $\mathbf{E}_t^{-1}$.
Intrinsics follow the standard pinhole form with focal lengths $(f_x, f_y)$ and principal point $(c_x, c_y)$.
The VAE has spatial stride $s$ dividing the pixel resolution $H \times W$ into the latent resolution $h \times w = (H/s) \times (W/s)$, with $s = 16$ throughout.
Latent cells are indexed by $(u, v) \in \{0, \dots, w-1\} \times \{0, \dots, h-1\}$ with pixel-centre homogeneous coordinates $[u + \tfrac{1}{2}, v + \tfrac{1}{2}, 1]^\top$.

\paragraph{Latent-resolution intrinsics.}
The latent-resolution intrinsic matrix used by the lifting and readout operations in the main paper is obtained from the pixel-resolution intrinsics $K$ by scaling both axes by the corresponding stride ratio,
\begin{equation}
K^{\ell} = \mathrm{diag}(w/W,\, h/H,\, 1)\, K.
\end{equation}
Both focal lengths and the principal point are rescaled by the same per-axis ratio, so that perspective projection remains consistent after the depth map is downsampled to latent resolution.

\paragraph{Pinhole back-projection.}
The lifting operator $\pi^{-1}$ maps a latent cell $(u, v)$ with depth $d = D(u, v)$ into a world point through ray-casting in camera coordinates followed by the camera-to-world transform,
\begin{equation}
\begin{aligned}
\tilde{\mathbf{u}} &= [u + \tfrac{1}{2},\, v + \tfrac{1}{2},\, 1]^\top, \\
\pi^{-1}(u, v, d;\, K^{\ell}, \mathbf{E})
&= \left.\mathbf{E}^{-1}\!\begin{bmatrix}
d\,(K^{\ell})^{-1}\tilde{\mathbf{u}} \\ 1
\end{bmatrix}\right|_{1:3}.
\end{aligned}
\end{equation}
where the trailing subscript denotes selection of the first three coordinates.

\paragraph{Projection onto the latent grid.}
For a memory point $\mathbf{p}_i$, its camera-space position at target view $t$ is the first three coordinates of $\mathbf{E}_t [\mathbf{p}_i^\top, 1]^\top$, which we denote $\mathbf{q}_i \in \mathbb{R}^3$.
Its position on the latent grid is then
\begin{equation}
\pi^{\ell}(\mathbf{q}_i) = \bigl(\lfloor x \rfloor, \lfloor y \rfloor\bigr), \qquad [x, y, 1]^\top = K^{\ell}\, \mathbf{q}_i / [\mathbf{q}_i]_z.
\end{equation}
The candidate set used in the main paper's readout operation is then
\begin{equation}
\Omega_t(u, v) = \bigl\{ i : \pi^{\ell}(\mathbf{q}_i) = (u, v),\; [\mathbf{q}_i]_z > 0 \bigr\},
\end{equation}
and the admissible cell set $\Lambda_t$ used in the memory update is defined on the same grid, restricted to cells with finite positive depth that lie outside the dynamic-object and sky masks.

\section{Additional Experimental Analysis}
\label{sec:app_experiment}

\begin{figure*}[!t]
    \centering
    \includegraphics[width=1.0\textwidth]{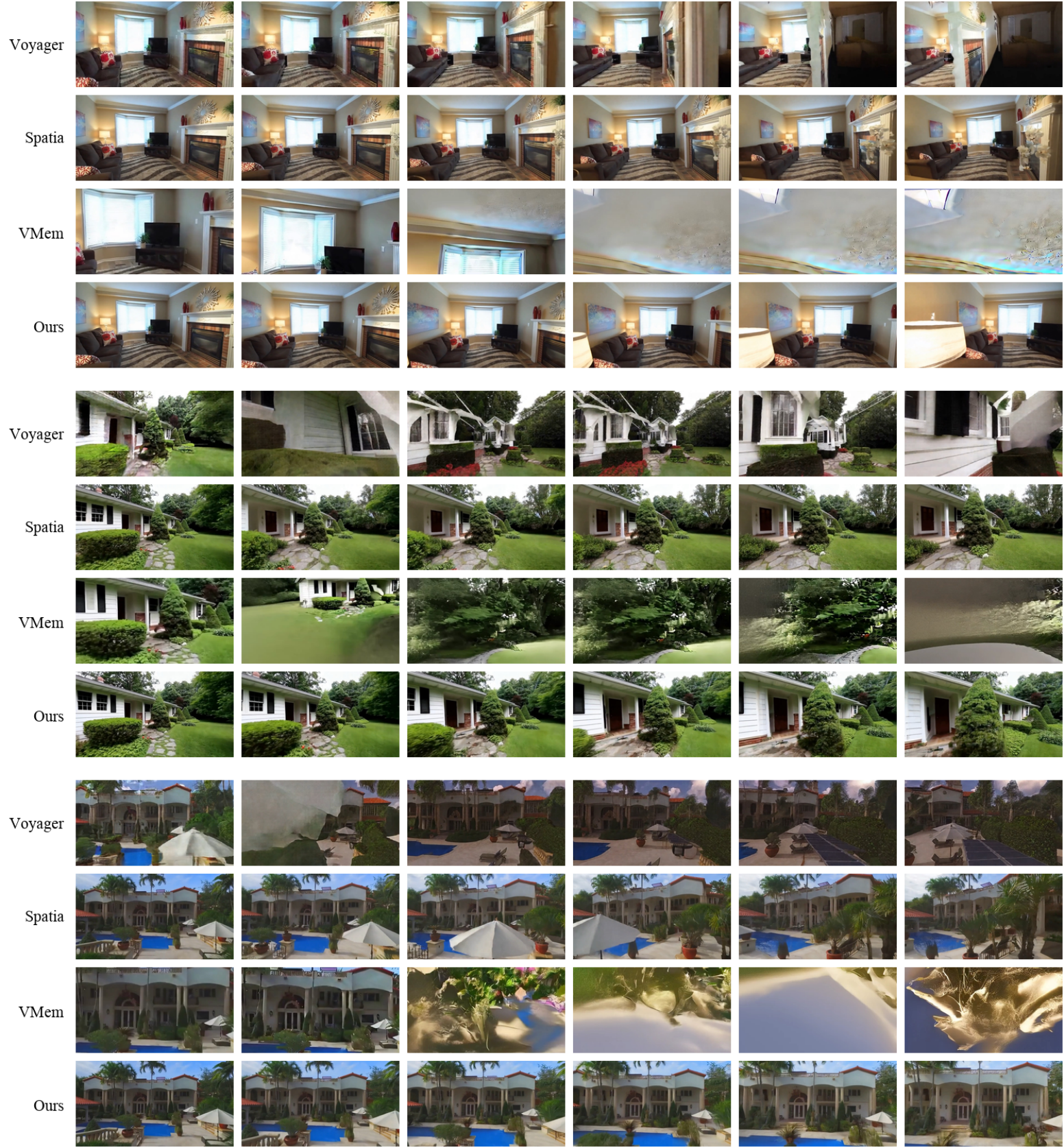}
    \caption{\textbf{Video comparison on RealEstate10K.}
    Each block shows one RealEstate10K trajectory, with rows corresponding to
    Voyager, Spatia, VMem, and \method{} (Ours), and columns showing uniformly
    sampled frames over time. Across indoor and outdoor scenes, \method{}
    preserves sharper structure and more stable appearance under camera motion,
    while baselines exhibit geometry drift and accumulated artifacts.}
    \label{fig:re10k_video}
\end{figure*}

In this section, we provide analysis that clarify \emph{why} latent spatial memory is efficient, \emph{how} it behaves as the rollout horizon grows, and \emph{what} the stored tokens carry compared with an RGB cache.

\paragraph{Cost of cache reads.}
Let $N$ denote the cache size at a given step.
Reading an RGB cache costs $\Theta(N \log N + HW) + \Theta(\Phi_\mathcal{E}(H, W))$ per conditioning step, where $\Phi_\mathcal{E}$ is the FLOP count of one VAE encoder pass.
Reading a latent cache costs $\Theta(N \log N + hw)$: the rasterisation term shrinks by the squared VAE stride, and the per-step encoder term disappears entirely because the readout is already in the backbone's input space.
The peak memory of the cache itself follows the ratio $s^2 \cdot (3 / C)$ between the RGB and latent representations, which together with the per-step pixel-resolution buffer accounts for the order-of-magnitude gap reported in the main paper.
The gap widens with the rollout horizon: the $N \log N$ sort begins to dominate the latent cache much later than the pixel cache, and the RGB baseline exhausts GPU memory on trajectories that \method{} completes comfortably.

\paragraph{Per-step timing breakdown.}
We decompose the per-step cost of the RGB pipeline into rasterisation, VAE encoder forward, and depth estimation plus back-projection.
On a $257$-frame rollout, rasterisation and the encoder together account for more than $98\%$ of the per-step RGB cost, and both are absent from the conditioning loop of \method{}, since a single latent-resolution projection replaces them.
The decoder, which the RGB pipeline calls at every conditioning step to produce the image on which the cache is rasterised, is invoked in \method{} only once per chunk to materialise output frames and to feed the depth and segmentation modules; it never appears on the per-step critical path.

\paragraph{Depth down-sampling for cache construction.}
The lifting operation requires the dense depth map and the latent grid to share a resolution, so depth must be down-sampled before lifting.
Because depth is piecewise smooth with discontinuities at silhouettes, this choice is not neutral: bilinear interpolation smears foreground over background at edges and can spawn phantom points; nearest-neighbour preserves edges but aliases fine structure; area pooling over-smooths occluding contours; median pooling preserves edges but is biased on slanted surfaces.
We evaluate all four with every other component fixed.
Table~\ref{tab:depth_down} reports the fraction of empty latent cells in the projected cache (``hole rate'').
Among the four down-sampling choices, bilinear interpolation gives the lowest hole rate at $42.53\%$.
We therefore adopt bilinear as the default, since it provides the best projected-cache coverage among the tested variants.

\begin{table}[t]
    \centering
    \caption{\textbf{Depth down-sampling for cache construction.} Evaluated on a held-out RealEstate10K split with every other component of \method{} kept fixed.}
    \label{tab:depth_down}
    \begin{tabular}{lc}
    \toprule
    Down-sampling $D \to D^{\ell}$ & Hole Rate $\downarrow$ \\
    \midrule
    Bilinear (default) & 42.53 \\
    Nearest-neighbour  & 47.78 \\
    Area pooling       & 53.72 \\
    Median pooling     & 52.22 \\
    \bottomrule
    \end{tabular}
\end{table}

\paragraph{What the cache stores.}
The latent tokens are expected to carry semantic and textural structure that three color channels cannot express.
We visualize this by projecting each per-point feature vector onto its top three principal components and coloring the memory by the resulting RGB.
Coherent semantic clusters such as walls, floors, windows, and furniture emerge that are not recoverable from an RGB point cloud built on the same frames.
This is the qualitative mechanism behind the main paper's ablation results, where swapping the latent cache for an RGB cache with the backbone and training recipe held fixed weakens both 3D and photometric consistency.

\section{Implementation Details}
\label{sec:app_implementation}

\paragraph{Backbone.}
The backbone is Wan2.2-TI2V-5B~\cite{wan2025wan}, whose VAE has spatial stride $s = 16$, temporal stride $4$, and channel count $C = 48$.
Each generation chunk covers a $9 \times 44 \times 80$ latent tensor corresponding to $33$ RGB frames at $704 \times 1280$.
The transformer has hidden dimension $3072$, feed-forward dimension $14336$, $24$ attention heads, $30$ blocks, text context $512$ tokens, and uses RMS Q/K normalisation together with cross-attention normalisation.
The VAE is frozen throughout.

\paragraph{Conditioning branch.}
The latent readout $\hat{\mathbf{z}}_t$ is injected through a ControlNet-style side branch whose layout mirrors the VACE~\cite{jiang2025vace} blocks of Wan2.2.
Segment-aware rotary positional encoding tags each frame as noisy target, clean preceding, or clean reference inside a single forward pass; at inference, the denoised latents of the previous chunk become the preceding frames of the next.

\paragraph{Training.}
Training has two flow-matching~\cite{lipman2022flow} stages on the target frames.
Stage one updates only the side branch at learning rate $10^{-5}$ with the backbone and VAE frozen.
Stage two unlocks rank-$32$ LoRA adapters~\cite{hu2021lora} on the $\{\texttt{q}, \texttt{k}, \texttt{v}, \texttt{o}\, \texttt{ffn}$ projections of every self-attention layer and optimises them at learning rate $10^{-4}$.
Optimisation uses AdamW ($\beta = (0.0, 0.999)$, weight decay $10^{-3}$), bfloat16 mixed precision under FSDP sharding, gradient checkpointing, and text-dropout probability $0.2$.
At inference we use the UniPC flow scheduler with $40$ steps and classifier-free guidance disabled; efficiency is reported on a single NVIDIA H100.

\begin{algorithm}[t]
\caption{One rollout of \method{}.}
\label{alg:latmem}
\begin{algorithmic}[1]
\REQUIRE initial frame $I_0$; camera trajectory $\{(\mathbf{E}_t, K_t)\}_{t=0}^{T}$ with $\mathbf{E}_0$ fixed to the world frame
\STATE $\mathbf{z}_0 \leftarrow \mathcal{E}(I_0)$
\STATE $D_0 \leftarrow \textsc{DepthAnything3}(I_0)$
\STATE $M_0 \leftarrow \textsc{SAM3}(\textsc{Qwen3-VL}(I_0)) \cup \textsc{sky}(I_0)$
\STATE $\mathcal{M} \leftarrow \mathrm{Lift}(\mathbf{z}_0, D_0, K_0, \mathbf{E}_0, \Lambda_0)$
\STATE $\tau \leftarrow 0$; $\mathcal{O} \leftarrow \{I_0\}$ \COMMENT{collected output frames}
\WHILE{$\tau < T$}
    \STATE sample latent chunk $W = \{\tau + 1, \dots, \tau + |W|\}$
    \FOR{$t \in W$}
        \STATE $\hat{\mathbf{z}}_t, \mathbf{m}_t \leftarrow$ readout of $\mathcal{M}$ at $(\mathbf{E}_t, K_t)$
    \ENDFOR
    \STATE $\mathcal{H}_W \leftarrow \{(\hat{\mathbf{z}}_t, \mathbf{m}_t) : t \in W\}$
    \STATE $\{\mathbf{z}_t\}_{t \in W} \leftarrow \mathrm{Backbone}(\mathcal{H}_W, \text{preceding}, \text{reference})$
    \FOR{$t \in W$}
        \STATE $I_t \leftarrow \mathcal{D}(\mathbf{z}_t)$; append $I_t$ to $\mathcal{O}$
        \STATE $D_t \leftarrow \textsc{DepthAnything3}(I_t)$
        \STATE $M_t \leftarrow \textsc{SAM3}(\textsc{Qwen3-VL}(I_t)) \cup \textsc{sky}(I_t)$
        \STATE $\tilde{\mathbf{z}}_t \leftarrow \mathcal{E}(I_t)$
        \STATE $\Delta\mathcal{M}_t \leftarrow \mathrm{Lift}(\tilde{\mathbf{z}}_t, D_t, K_t, \mathbf{E}_t, \Lambda_t)$
        \STATE $\mathcal{M} \leftarrow \mathcal{M} \cup \Delta\mathcal{M}_t$
    \ENDFOR
    \STATE $\tau \leftarrow \tau + |W|$
\ENDWHILE
\STATE \RETURN $\mathcal{O}$
\end{algorithmic}
\end{algorithm}

\paragraph{Data.}
Training clips are drawn from RealEstate10K.
Each clip is processed by ViPE~\cite{huang2025vipe} for per-frame depth, intrinsics, and extrinsics, and by a Qwen3-VL-2B~\cite{yang2025qwen3} entity extractor followed by SAM3~\cite{carion2025sam3segmentconcepts} for foreground-dynamic and sky masks.
Cells inside the mask are excluded from $\Lambda_t$ during memory update so that only geometry compatible with the rigid-scene assumption enters the persistent memory.
Frames, latents, depth, and camera parameters are stored in an LMDB-backed dataset, so no re-encoding occurs during training.
Rollouts are produced in chunks of nine latent frames with one-frame overlap to preserve temporal continuity, and the cache is grown once per chunk by re-encoding the decoded frames and lifting them into the memory.
Algorithm~\ref{alg:latmem} summarises one complete rollout.

\subsection{Human Evaluation}
\label{sec:exp:user-study}

To complement automatic metrics, we design a blinded human study that measures the properties most directly affected by persistent spatial memory. Each trial presents the input image, the prescribed camera trajectory, and a generated clip, including a short segment that returns to an earlier viewpoint. Raters score each criterion independently on a scale from $0$ (unusable) to $5$ (excellent). 

The protocol follows questions in Table~\ref{tab:user-study-metrics}, which separates camera following from other dimensions, and makes the revisit consistency segment explicit so that long-horizon memory is evaluated rather than only local frame quality.

\begin{table}[H]
    \centering
    \caption{\textbf{User-study criteria.} All criteria use a $0$--$5$ scale, where higher is better.}
    \begin{tabular}{>{\raggedright\arraybackslash}p{0.29\linewidth}p{0.62\linewidth}}
    \toprule
    Criterion & Question \\
    \midrule
    Scene Revisit Consistency & After leaving and returning to an earlier region, do the same surfaces and objects reappear with the same layout, identity, and spatial relationships? \\
    Camera Following & Does the generated motion follow the prescribed camera path without unintended camera drift? \\
    Appearance Fidelity & Does the same surface retain its color, texture, illumination, and fine detail across viewpoints and when revisited? \\
    Temporal Smoothness & Are transitions between adjacent frames smooth and free of flicker, ghosting, or abrupt changes? \\
    Overall Quality & How convincing and usable is the clip in general, considering realism, detail, consistency, and visible artifacts? \\
    \bottomrule
    \end{tabular}
    \label{tab:user-study-metrics}
\end{table}

\begin{figure}[H]
    \centering
    \includegraphics[width=\linewidth]{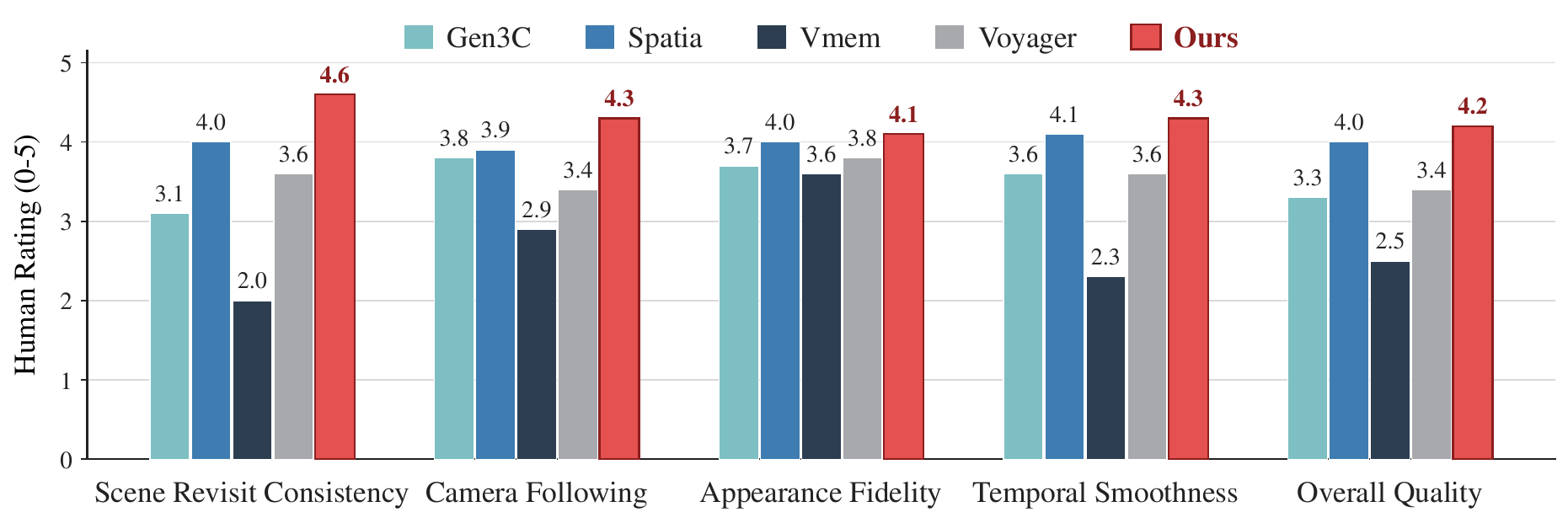}
    \caption{\textbf{Human-study ratings.} Mean ratings for the five criteria on a 0--5 scale; higher is better.}
    \label{fig:user-study}
\end{figure}

We sample scenes randomly from RealEstate10K and WorldScore, and evaluate all methods under identical inputs and trajectories. Clips are anonymized and shown in a randomized order. We target 25 raters and 30 scenes, with each clip receiving at least five independent ratings. As shown in Figure~\ref{fig:user-study}, \method{} achieves the highest mean score on all five dimensions, including 4.6 for scene revisit consistency, 4.3 for camera following, 4.1 for appearance fidelity, 4.3 for temporal smoothness, and 4.2 for overall quality.

\end{document}